\documentclass{article}
\usepackage{spconf,amsmath,graphicx,amssymb}
\usepackage{booktabs}
\usepackage{multirow}
\usepackage{graphicx}
\usepackage{xcolor}
\usepackage{url}
\usepackage{cite}
\usepackage{hyperref}       

\title{Unsupervised Point Cloud Pre-Training via Contrasting and Clustering}
\name{Guofeng Mei$^{1}$, Xiaoshui Huang$^{2}$, Juan Liu$^{3}$, Jian Zhang$^{1}$, Qiang Wu$^{1}$}
\address{
$^{1}$University of Technology Sydney \quad
$^{2}$Shanghai AI Laboratory \quad
$^{3}$Beijing Institute of Technology
}

\begin{document}
%
\maketitle
\begin{abstract}
Annotating large-scale point clouds is highly time-consuming and often infeasible for many complex real-world tasks. Point cloud pre-training has therefore become a promising strategy for learning discriminative representations without labeled data. In this paper, we propose a general unsupervised pre-training framework, termed \emph{ConClu}, which jointly integrates contrasting and clustering. The contrasting objective maximizes the similarity between feature representations extracted from two augmented views of the same point cloud, while the clustering objective simultaneously partitions the data and enforces consistency between cluster assignments across augmentations. Experimental results on multiple downstream tasks show that our method outperforms state-of-the-art approaches, demonstrating the effectiveness of the proposed framework. Code is available at \url{https://github.com/gfmei/conclu}.
\end{abstract}

\section{Introduction}
With recent advances in depth sensing technology, acquiring point clouds has become increasingly convenient and affordable, leading to their widespread use in a variety of applications~\cite{xu2020geometry,mei2022partial}. Learning discriminative and transferable feature representations from point clouds is a fundamental problem in 3D shape understanding, as it enables effective training for downstream tasks such as object detection and tracking, segmentation, reconstruction, and classification~\cite{guo2020deep}. However, such success often relies on large amounts of manually annotated data. Annotating point clouds is particularly challenging for several reasons. First, their sparse, low-resolution, and irregular spatial structure makes accurate labeling difficult~\cite{wang2020unsupervised}. Second, the large number of points in each sample substantially increases annotation cost and reduces efficiency~\cite{wang2020unsupervised}. To alleviate this burden, many efforts have been devoted to reducing the dependence on labeled data, among which unsupervised pre-training has shown considerable promise.

A number of studies have explored unsupervised pre-training methods for learning effective point cloud representations~\cite{guo2020deep}. Existing approaches can generally be divided into two categories: generative and discriminative methods~\cite{grill2020bootstrap}. Generative methods include self-reconstruction and auto-encoding approaches~\cite{yang2018foldingnet}, generative adversarial networks (GANs)~\cite{sarmad2019rl}, and auto-regressive models. These methods typically encode an input point cloud into a global latent representation~\cite{shi2020unsupervised}, or into a latent distribution in the variational setting~\cite{hassani2019unsupervised}, and then reconstruct the input through a decoder. Although such methods are effective at capturing high-level structural information, many of them implicitly assume that objects within the same category share a canonical pose~\cite{sanghi2020info3d}. As a result, they are often sensitive to geometric transformations such as rotation and translation.

In contrast, discriminative approaches learn representations by predicting or distinguishing among different data augmentations. These methods tend to better preserve semantic information and have recently demonstrated strong performance in unsupervised point cloud representation learning. Among them, contrastive learning methods~\cite{sanghi2020info3d} have achieved particularly promising results. By contrasting different augmented views of the same sample, they can learn representations that are robust to transformations and preserve semantic consistency. The key idea of contrastive learning is to bring representations of positive pairs closer while pushing those of negative pairs farther apart~\cite{chen2020simple}. However, high performance often depends on a large number of negative samples and is highly sensitive to their selection~\cite{grill2020bootstrap,chen2020simple}. Consequently, these methods are usually computationally expensive, often requiring large batch sizes, memory banks, or carefully designed negative-sample mining strategies~\cite{grill2020bootstrap}. Although methods such as BYOL~\cite{grill2020bootstrap} and SimSiam~\cite{chen2020simple} have shown that competitive self-supervised learning is possible without explicit negative pairs in 2D vision tasks, the issue of representation collapse remains a concern~\cite{chen2020simple}. Therefore, reducing the reliance on negative samples while avoiding collapse remains an open challenge for unsupervised point cloud representation learning.

To address these issues, we propose an unsupervised point cloud pre-training framework that achieves state-of-the-art performance without using negative pairs. Our method jointly performs contrasting and clustering, and directly maximizes the similarity between two global features extracted from two augmented views of the same point cloud. Specifically, the contrasting component is inspired by the SimSiam framework~\cite{chen2021exploring}, while the clustering component is introduced to further prevent the network from collapsing to a constant solution. In our framework, a stop-gradient operation and a constraint that encourages point clouds to be distributed uniformly across clusters are employed to avoid degenerate solutions. The stop-gradient operation blocks gradients from propagating through one branch during backpropagation, thereby stabilizing training and helping prevent collapse.

\section{The Proposed Method}
\subsection{Overview}
Consider a 3D point cloud $P=\{p_i\in\mathbb{R}^3 \mid i=1,2,\ldots,N\}$ consisting of $N$ points, where each point $p_i\in\mathbb{R}^3$ is represented by its 3D coordinates. Our goal is to pre-train a encoder $f_\phi$ (e.g., PointNet) for feature extraction with parameters $\phi$ in an unsupervised manner.

\begin{figure}[t]
    \centering
    \includegraphics[width=1.0\columnwidth]{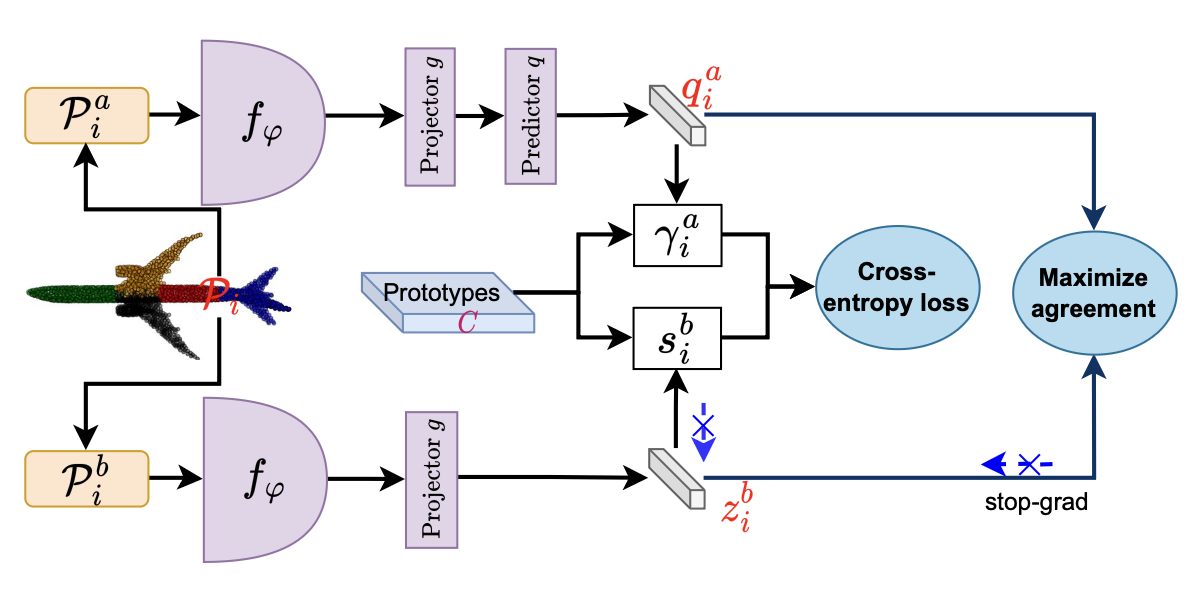}
    \vspace{-3mm}
    \caption{The illustration of our unsupervised approach.}
    \label{fig:framework}
\end{figure}

As illustrated in Fig.~\ref{fig:framework}, our framework takes as input two randomly augmented views, $P_i^a$ and $P_i^b$, generated from the same point cloud $P_i$. These two views are processed by a neural network composed of a shared encoder backbone $f_\phi$, a max-pooling operator $\rho$, and a projection MLP head $g$~\cite{chen2020simple}. The encoder $f_\phi$ and the projection head $g$ share weights across the two branches. A prediction MLP head, denoted by $q$~\cite{grill2020bootstrap}, is then used to transform the output of one branch so that it can be matched to the representation of the other branch. Notably, the predictor is applied only to one branch, resulting in an asymmetric architecture~\cite{grill2020bootstrap}.

Denote the two output vectors as
\begin{equation}
q_i^a \triangleq q\bigl(g(\rho(F_i^a))\bigr), \qquad z_i^b \triangleq g(\rho(F_i^b)),
\end{equation}
where $F_i^a=f_\phi(P_i^a)$ and $F_i^b=f_\phi(P_i^b)$. Following~\cite{chen2020simple}, we apply a stop-gradient (sg) operation to $z_i^b$ to avoid collapse to a trivial constant solution in the absence of negative samples. The contrasting module maximizes the agreement between $q_i^a$ and $z_i^b$. In parallel, the clustering module generates the pseudo-label $s_i^b$ and the predicted label $\gamma_i^a$ by assigning $z_i^b$ and $q_i^a$, respectively, to a set of learned prototype vectors $C$. The prediction objective is then defined by minimizing the cross-entropy loss between $s_i^b$ and $\gamma_i^a$. By symmetry, exchanging $P_i^a$ and $P_i^b$ yields $q_i^b$, $z_i^a$, $s_i^a$, and $\gamma_i^b$.

\subsection{Contrasting module}
The contrasting module is formulated by maximizing the agreement between $q_i^a$ and $z_i^b$. Specifically, we define the following mean squared error between the $\ell_2$-normalized prediction $q_i^a$ and projection $z_i^b$ to measure their agreement:
\begin{equation}
D(q_i^a,z_i^b) \triangleq \left\|\frac{q_i^a}{\|q_i^a\|_2}-\frac{z_i^b}{\|z_i^b\|_2}\right\|_2^2
=2-\frac{2 (q_i^a)^\top z_i^b}{\|q_i^a\|_2\cdot \|z_i^b\|_2}.
\label{eq:1}
\end{equation}
This is equivalent to the negative cosine similarity, up to a scale of 2. To prevent collapsing solutions, a stop-gradient operation is applied on $z_i^b$ by modifying Eq.~\eqref{eq:1} as
\begin{equation}
D\bigl(q_i^a,\operatorname{sg}(z_i^b)\bigr).
\label{eq:2}
\end{equation}
Stop-gradient operation stops the accumulated gradient of the inputs from flowing through this operator in the backward direction. This means that $z_i^b$ is treated as a constant vector in this term. Following~\cite{chen2020simple}, we define a symmetrized loss as
\begin{equation}
\mathcal{L}_{\text{con}} = D\bigl(q_i^a,\operatorname{sg}(z_i^b)\bigr) + D\bigl(q_i^b,\operatorname{sg}(z_i^a)\bigr).
\label{eq:3}
\end{equation}
The minimum possible value of Eq.~\eqref{eq:3} is 0. The encoder on $q_i^a$ receives no gradient from $z_i^b$ in the first term, but it receives gradients from $q_i^b$ in the second term (and vice versa for $q_i^b$). Without adopting the stop-gradient, despite a zero loss during the training process, the representations learned are useless, as all point clouds tend to get the same representation~\cite{chen2020simple}, i.e., the model collapses to a constant mapping.

\subsection{Clustering module}

Given a mini-batch of $B$ feature vectors $Q^k = [q_1^k, q_2^k, \cdots, q_B^k],$ $k \in \{a,b\}$ with $q_i^k \triangleq q\bigl(g(\rho(f_{\phi}(\mathcal{P}_i^k)))\bigr),$
we map each feature $q_i^k$ to a set of $J$ learnable clustering prototypes
\[
C = [c_1, c_2, \dots, c_J].
\]
The probability $\gamma_{ij}^k \in \gamma^k$ that $q_i^k$ belongs to the $j$-th cluster is computed by applying a softmax over the cosine similarities between $q_i^k$ and all prototypes in $C$, i.e.,
\begin{equation}
\gamma_{ij}^k =
\frac{\exp\bigl(\cos(c_j, q_i^k)/\tau\bigr)}
{\sum_{l=1}^{J} \exp\bigl(\cos(c_l, q_i^k)/\tau\bigr)},
\label{eq:cluster_prob}
\end{equation}
where $\tau = 10^{-2}$ is a temperature parameter.

We then compute the codes
\[
S^k = \{s_{ij}^k\}_{i,j}
\]
using the prototypes $C$ such that all samples in a mini-batch are approximately equally partitioned among the prototypes. This equipartition constraint encourages distinct codes for different point clouds within a batch, thereby preventing collapse. Intuitively, the prototypes $C$ are shared across different mini-batches. We denote the pseudo-labels by
\[
S^k = [s_1^k, s_2^k, \dots, s_B^k],
\qquad
s_i^k = \{s_{ij}^k\}_{j=1}^{J}, \qquad k \in \{a,b\}.
\]
The pseudo-labels are obtained by maximizing the similarity between the features and the prototypes. Specifically, $S^a$ is optimized by solving
\begin{equation}
\min_{S^a} \sum_{i=1}^{B} \sum_{j=1}^{J}
s_{ij}^a \, \mathcal{D}\bigl(c_j, \operatorname{sg}(z_i^b)\bigr),
\label{eq:ot_obj}
\end{equation}
subject to
\begin{equation}
S^a \mathbf{1}_{J} = \frac{1}{B}\mathbf{1}_{B},
\qquad
(S^a)^{\top}\mathbf{1}_{B} = \frac{1}{J}\mathbf{1}_{J},
\label{eq:ot_constraint}
\end{equation}
where $\mathbf{1}_t$ ($t=B,J$) denotes the all-ones vector of dimension $t$.
Similarly, $S^b$ can be optimized in the same manner. These constraints ensure that, on average, each prototype is assigned to at least $B/J$ samples within a mini-batch. The objective in Eq.~\eqref{eq:ot_obj} is an instance of the optimal transport~\cite{cortes1995support} problem, which can be efficiently solved using the Sinkhorn-Knopp algorithm~\cite{cortes1995support}.
To further avoid the degenerate case where all prototypes collapse to the same vector, we introduce an orthogonal regularization term:
\begin{equation}
\mathcal{L}_{\text{orth}}
=
\left\| \Gamma^{\top}\Gamma - I \right\|_{1},
\label{eq:orth}
\end{equation}
where $\Gamma =
\left[
\frac{c_1}{\|c_1\|_2},
\frac{c_2}{\|c_2\|_2},
\dots,
\frac{c_J}{\|c_J\|_2}
\right].$
By applying this loss over all point clouds and both augmented views, the clustering loss is defined as
\begin{equation}
\mathcal{L}_{\text{clu}}
=
\frac{1}{B}\sum_{i=1}^{B}\sum_{j=1}^{J}
\left(
s_{ij}^a \log \gamma_{ij}^b
+
s_{ij}^b \log \gamma_{ij}^a
\right)
+
\mathcal{L}_{\text{orth}}.
\label{eq:clu}
\end{equation}
Finally, the overall objective is given by
\begin{equation}
\mathcal{L}_{\text{total}}
=
\mathcal{L}_{\text{con}}
+
\mathcal{L}_{\text{clu}}.
\label{eq:total}
\end{equation}

\section{Experiments}
In this section, we present the implementation details, the setup of pre-training and the downstream fine-tuning. Then, we explore the performance of the results for object classification and 3D part segmentation.

\subsection{Implementation details}
Our implementation is built on the Pytorch~\cite{paszke2019pytorch} library. We used AdamW~\cite{loshchilov2016sgdr} as the default optimizer with the learning rate of 0.001. The batch size was set to 32, and the learning rate was delayed by 0.7 every 20 epochs. We trained our model for 200 epochs since the loss function stabilized at around the 200th epoch. Our models were trained on two Tesla V100-PCI-E-32G GPUs.

\paragraph*{Data augmentation}
The stochastic data augmentation module generates two correlated views, $P_i^a$ and $P_i^b$, from each input point cloud $P_i$, which are treated as a positive pair. In our implementation, three augmentations are applied sequentially: random cropping followed by point sampling, random rotation, and random jittering.

\begin{itemize}
    \item \textbf{Random Cropping:} For each point cloud, we sample a half-space from the unit sphere $S^2=\{x\in\mathbb{R}^3 \mid \|x\|=1\},$
    with a random direction $s\in S^2$, and shift it such that approximately 85\% of the points are preserved.
    
    \item \textbf{Random Rotation:} Each point cloud is randomly rotated around the vertical axis, with the rotation angle uniformly sampled from $[-5.0^\circ, 5.0^\circ]$.
    
    \item \textbf{Random Jittering:} Each point is randomly perturbed by Gaussian noise sampled from $\mathcal{N}(0, 0.01)$ and clipped to $[-0.025, 0.025]$ along each axis.
\end{itemize}

\paragraph*{Point segmentation backbone}
We directly compare our implementation against Jigsaw3D~\cite{sauder2019self} and OcCo~\cite{wang2020unsupervised} by extracting the representation. Following Jigsaw3D, two 3D backbone networks, PointNet~\cite{qi2017pointnet} and Dynamic Graph CNN (DGCNN)~\cite{wang2019dynamic}, are implemented for fair comparisons. Though we choose PointNet and DGCNN as backbones, our approach is flexible with any neural network architecture designed for point cloud classification. In all cases, our latent dimension is set to 1024 to match prior works.

\subsection{Pre-training setup}
For all experiments, we pre-train the backbone networks on ModelNet40~\cite{sharma2016vconv} using the proposed pretext task. ModelNet40 contains 12{,}331 CAD models from 40 object categories, split into 9{,}843 training samples and 2{,}468 testing samples. Following~\cite{huang2021spatio}, we randomly sample 2{,}048 points from the surface of each model to form a point cloud, and then apply data augmentation to obtain different views. The pre-training set is constructed from the training split of ModelNet40. The number of clusters is set to $J=32$, and the batch size is set to 128.
The projector $g(\cdot)$ and predictor $q(\cdot)$ are designed as follows:
\begin{itemize}
    \item \textbf{Projection MLP:} The projection head $g$ consists of three fully connected layers, each followed by batch normalization. The hidden dimension is 1024, and the final output dimension is 256. LeakyReLU is applied after each layer except the last.
    \item \textbf{Prediction MLP:} The predictor $q$ consists of a linear layer with output dimension 512, followed by batch normalization and LeakyReLU, and then a final linear layer with output dimension 256.
\end{itemize}

\subsection{Fine-tuning setup}
\noindent\textbf{Object classification}
We evaluate the shape understanding capability of our unsupervised learning model on two object classification benchmarks, ModelNet40~\cite{sharma2016vconv} and ModelNet10~\cite{sharma2016vconv}. ModelNet10 contains 4{,}899 pre-aligned shapes from 10 categories, with 3{,}991 samples for training and 908 for testing. Following the standard protocol in prior work~\cite{wang2020unsupervised}, we adopt a linear SVM classifier to assess the quality of the learned representations. Specifically, the SVM is trained on the global features extracted from the training split of ModelNet40/10. Each point cloud is formed by randomly sampling 2{,}048 points from the surface of each shape.
We compare ConClu with a range of previous generative and contrastive methods, as well as PointNet- and DGCNN-based models trained with different pretext tasks. The classification results on the test sets are summarized in Table~\ref{tab:classification}. ConClu consistently outperforms all competing methods that use the same backbone architecture. On ModelNet40, our PointNet-based model achieves a classification accuracy of 89.8\%, outperforming both the second-best generative method, OcCo~\cite{wang2020unsupervised}, which achieves 88.7\%, and the contrastive method STRL$^{\star}$~\cite{huang2021spatio}, which achieves 88.3\% despite being pre-trained on the larger ShapeNet~\cite{yi2016scalable} dataset. Notably, the linear SVM performance of our method even surpasses that of fully supervised PointNet, which achieves 89.2\% test accuracy when trained from random initialization. Our DGCNN-based model achieves 91.6\% test accuracy on ModelNet40, exceeding the second-best method, STRL$^{\star}$, by 0.7\%. Our method also delivers competitive results on ModelNet10, further demonstrating its effectiveness.

\begin{table}[t]
\centering
\caption{Comparisons of the classification accuracy (\%) of our method against state-of-the-art unsupervised point cloud pre-training methods on ModelNet40 and ModelNet10. $\star$ indicates that the model is pre-trained on ShapeNet dataset.}
\label{tab:classification}
\tabcolsep 6pt
{%
\begin{tabular}{lccc}
\toprule
Method & Year & ModelNet40 & ModelNet10 \\
\midrule
LGAN~\cite{achlioptas2018learning} & 2018 & 87.3 & 92.2 \\
LGAN$^{\star}$ & 2018 & 85.7 & 95.3 \\
FoldingNet~\cite{yang2018foldingnet} & 2018 & 84.4 & 91.9 \\
FoldingNet$^{\star}$ & 2018 & 88.4 & 94.4 \\
MRTNet$^{\star}$~\cite{gadelha2018multiresolution} & 2018 & 86.4 & -- \\
ContrastNet~\cite{zhang2019unsupervised} & 2019 & 86.8 & 93.8 \\
MAP-VAE$^{\star}$~\cite{han2019multi} & 2019 & 90.2 & 94.8 \\
\midrule
\multicolumn{4}{l}{PointNet}\\
Jigsaw3D~\cite{sauder2019self} & 2019 & 87.5 & 91.3 \\
Jigsaw3D$^{\star}$~\cite{sauder2019self} & 2019 & 87.3 & 91.6 \\
Rotation3D$^{\star}$~\cite{poursaeed2020self} & 2020 & 88.6 & -- \\
 OcCo~\cite{wang2020unsupervised} & 2021 & 88.7 & 91.4 \\
STRL$^{\star}$~\cite{huang2021spatio} & 2021 & 88.3 & -- \\
ConClu (ours) & -- & \textbf{89.8} & \textbf{93.3} \\
\midrule
\multicolumn{4}{l}{DGCNN}\\
Jigsaw3D~\cite{sauder2019self} & 2019 & 87.8 & 92.6 \\
Jigsaw3D$^{\star}$~\cite{sauder2019self} & 2019 & 90.6 & 94.5 \\
Rotation3D$^{\star}$~\cite{poursaeed2020self} & 2020 & 90.8 & -- \\
OcCo~\cite{wang2020unsupervised} & 2021 & 89.2 & 92.7 \\
STRL$^{\star}$~\cite{huang2021spatio} & 2021 & 90.9 & -- \\
ConClu (ours) & -- & \textbf{91.6} & \textbf{95.0} \\
\bottomrule
\end{tabular}}
\end{table}

\noindent\textbf{Part segmentation}
We further evaluate the transferability of the learned representations on 3D part segmentation. Following prior work, we evaluate two backbone networks, PointNet and DGCNN, on the ShapeNetPart~\cite{yi2016scalable} benchmark. ShapeNetPart contains 16{,}881 objects from 16 categories with a total of 50 part labels, and each object consists of 2{,}048 points. During testing, we use the same post-processing strategy as in~\cite{qi2017pointnet}. We report overall accuracy (OA) and mean intersection over union (mIoU) as the evaluation metrics.
As shown in Table~\ref{tab:seg}, our method consistently outperforms random initialization and previous unsupervised pre-training methods, including Jigsaw~\cite{sauder2019self} and OcCo~\cite{wang2020unsupervised}, in terms of both OA and mIoU. For PointNet, our pre-training improves performance over random initialization by 0.8\% in OA and 1.5\% in mIoU. For DGCNN, our method achieves 94.7\% OA and 85.4\% mIoU, exceeding random initialization by 2.5\% in OA and 1.0\% in mIoU. It also slightly outperforms the second-best method, OcCo, by 0.3\% in OA and 0.4\% in mIoU. These results demonstrate that the proposed framework learns transferable point-wise representations that are effective for fine-grained 3D part segmentation.

\begin{table}[t]
\centering
\caption{Part segmentation results.}
\label{tab:seg}
\tabcolsep 10pt
\begin{tabular}{llcc}
\toprule
Encoder & Method & OA (\%) & mIoU (\%) \\
\midrule
\multirow{4}{*}{PointNet}
& Random & 92.8 & 82.2 \\
& Jigsaw~\cite{sauder2019self} & 93.1 & 82.2 \\
& OcCo~\cite{wang2020unsupervised} & 93.4 & 83.4 \\
& Ours ConClu & \textbf{93.6} & \textbf{83.7} \\
\midrule
\multirow{4}{*}{DGCNN}
& Random & 92.2 & 84.4 \\
& Jigsaw~\cite{sauder2019self} & 92.7 & 84.3 \\
& OcCo~\cite{wang2020unsupervised} & 94.4 & 85.0 \\
& Ours ConClu & \textbf{94.7} & \textbf{85.4} \\
\bottomrule
\end{tabular}
\end{table}

\subsection{Ablation study}
The ablation results on ModelNet40 are summarized in Tab.~\ref{tab:ablation} (MN40). Using only the contrasting module yields classification accuracies of 91.2\% for DGCNN and 88.7\% for PointNet. Incorporating the clustering module further improves performance, leading to gains of 1.1\% and 0.4\% for PointNet and DGCNN, respectively. A similar trend can be observed on ModelNet10, as shown in Tab.~\ref{tab:ablation} (MN10). When only the contrasting module is used, DGCNN achieves an accuracy of 93.8\%, while PointNet attains 92.4\%. By combining contrasting with clustering, the accuracy is further improved by 0.9\% for PointNet and 1.2\% for DGCNN. These results clearly demonstrate the effectiveness of our method.

\begin{table}[t]
\centering
\tabcolsep 6pt
\caption{Ablation study of our method. We report the classification accuracy (\%) on ModelNet40/10. ($\mathcal{L}_{\text{global}}$: instance-level contrasting, $\mathcal{L}_{\text{local}}$: point-level clustering.)}
\label{tab:ablation}
\begin{tabular}{lccccc}
\toprule
Model & Contrasting & Clustering & MN40 & MN10 \\
\midrule
\multirow{3}{*}{PointNet}
& \checkmark &  & 88.7 & 92.4 \\
&  & \checkmark & 85.3 & 92.1 \\
& \checkmark & \checkmark & \textbf{89.8} & \textbf{93.3} \\
\midrule
\multirow{3}{*}{DGCNN}
& \checkmark &  & 91.2 & 93.8 \\
&  & \checkmark & 90.3 & 93.5 \\
& \checkmark & \checkmark & \textbf{91.6} & \textbf{95.0} \\
\bottomrule
\end{tabular}
\end{table}

\section{Conclusion}
This paper proposed a general unsupervised point cloud pre-training framework. Our method demonstrates strong transferability of the learned representations across downstream tasks, including 3D object classification and part segmentation. Moreover, our framework is independent of any specific neural network architecture, making it a generic and effective module for feature learning from raw point clouds and for improving the performance of a wide range of 3D models.

\bibliographystyle{IEEEbib}
\bibliography{main}

\end{document}